# Focusing on Shadows for Predicting Heightmaps from Single Remotely Sensed RGB Images with Deep Learning

**Savvas Karatsiolis**
CYENS Center of Excellence,
Nicosia, Cyprus
s.karatsiolis@cyens.org.cy

**Andreas Kamilaris**
CYENS Center of Excellence,
Nicosia, Cyprus
a.kamilaris@cyens.org.cy

**Abstract**

Estimating the heightmaps of buildings and vegetation in single remotely sensed images is a challenging problem. Effective solutions to this problem can comprise the stepping stone for solving complex and demanding problems that require 3D information of aerial imagery in the remote sensing discipline, which might be expensive or not feasible to require. We propose a task-focused Deep Learning (DL) model that takes advantage of the shadow map of a remotely sensed image to calculate its heightmap. The shadow is computed efficiently and does not add significant computation complexity. The model is trained with aerial images and their Lidar measurements, achieving superior performance on the task. We validate the model with a dataset covering a large area of Manchester, UK, as well as the 2018 IEEE GRSS Data Fusion Contest Lidar dataset. Our work suggests that the proposed DL architecture and the technique of injecting shadows' information into the model are valuable for improving the heightmap estimation task for single remotely sensed imagery.

1. Introduction

Aerial images are widely used in geographic information systems (GIS) for a plethora of interesting tasks like urban monitoring and planning [1–3], agricultural development [4], landscape change detection [5–7], disaster mitigation planning and recovery [8], as well as aviation [9,10]. However, these images are primarily two-dimensional (2D) and constitute a poor source of three-dimensional (3D) information, a fact that hinders the adequate understanding of vertical geometric shapes and relations within a scene. Ancillary 3D information improves the performance of many GIS tasks and facilitates the development of tasks that require geometric analysis of the scene, such as digital twins for smart cities [11] and forest mapping [12]. In such cases, the most popular type of this complementary 3D information is the form of a Digital Surface Model (DSM). The DSM is often obtained with a Light Detection and Ranging Laser Scanner (Lidar) or an Interferometric Synthetic-

Aperture Radar (InSAR), a Structure from Motion (SfM) methodology [13], or by using stereo image pairs [14]. Structure from motion is a technique for estimating 3D structures from 2D image sequences. The main disadvantages of SfM include the possible deformation of the modeled topography, its over-smoothing effect, the necessity for optimal conditions during data acquisition and the necessity of a ground control point [15]. Using stereo image pairs to infer 3D information from aerial imagery is also costly and challenging [15]. Like SfM, DSM estimation by stereo image pairs requires difficult and sophisticated acquisition techniques, precise image pairing, and relies on triangulation from pairs of consecutive views. Lidar sensors have lately become affordable to use and can provide accurate height estimations [16]. At the same time, Lidar sensors suffer from poor performance when reflective surfaces (like water) occur in an image, and they sometimes return values that are irrationally high or incorrect, especially in cases of multiple reflections in a complex scene. Besides these flaws, Lidar is a commonly used technology for obtaining the heightmap of a scene. A substantial non-technical disadvantage of Lidar is the high cost associated with obtaining high-resolution aerial images of the sceneries of interest. Therefore, predicting the heightmap from a single image is a very compelling idea that has not been explored thoroughly by the remote sensing community. The reason for limited work on this topic stems from the fact that height estimation from a single image and monocular vision in general is an ill-posed problem: infinite possible heightmaps may correspond to a single image. This means that different height configurations of a scene may have the same remotely sensed image. Moreover, remotely sensed images frequently pose scale ambiguities that make inference on geometric relations very hard. Consequently, mapping 2D pixel-associated intensities to pixel real-world height values is a challenging task.

In contrast to the remote sensing research community, the computer vision (CV) community shows a significant interest in depth estimation from a single image. Depth perception is known to improve computer vision tasks like semantic segmentation [17,18], human pose estimation [19], and image recognition [20–22] in an analogous manner that height estimation is known to improve remote sensing tasks. Methods like stereo vision pairing, SfM, and various feature transfer strategies [23] were used for depth prediction before the successful application of Deep Learning (DL) on the task. All these methods require expensive and precise data (pre)processing to deliver quality results. Deep Learning simplifies the process while achieving better performance. Eigen et al. [24] use a multiscale architecture to predict the depth map from a single image. The first component of the architecture is based on the AlexNet DL architecture [22] and produces a coarse depth estimation refined by an additional processing stage. Laina et al. [25] introduce residual blocks into their DL model and use the reverse Huber loss for optimizing the depth prediction. Alhashmin and Wonka [26] use transfer learning from a DenseNet model [27] pre-trained on ImageNet [28], which they connect to a decoder using multiple skip connections. By applying multiple losses between the ground truth and the prediction ($L_1$ loss between predicted and ground truth values, $L_1$ loss between predicted map gradient and ground truth map gradient and Structural Similarity between the actual and the predicted map), they achieve the state-of-the-art depth estimation from a single image. The interest of the CV community on depth estimation originates from the need for better navigation for autonomous agents, space geometry perception and scene understanding, especially in the research fields of robotics and the autonomous cars' industry. Specifically, regarding monocular depth estimation, AdaBins [29] achieves state-of-the-art performance on KITTI [30] and NYU-Depth v2 [31] datasets by using adaptive bins for depth estimation. Mahjourian et al. [32] use a feature-metric loss for self-supervised learning of depth and ego-motion. Koutilya et al. [33] combine synthetic and real data for unsupervised geometry estimation through a generative adversarial network (GAN) that they call SharinGAN, which maps both real and synthetic images to a shared domain. SharinGAN achieves state-of-the-art performance on the KITTI dataset.

The main approaches used by researchers in aerial image height estimation based on DL involve: a) training with additional data, b) tackling auxiliary tasks in parallel to depth estimation, c) using deeper models with skip connections between layers, and c) using generative models like GANs with conditional settings. Alidoost et al. [34] apply a knowledge-based 3D building reconstruction by incorporating additional structural information regarding the buildings in the image, like lines from the structures' outline. Mou and Zhu [35] propose an encoder-decoder convolutional architecture called IM2HEIGT that uses a single but provenly functional skip connection from the first residual block to the second last block. They argue that both the use of the residual blocks and the skip connection contribute significantly to the model's performance. The advantages of using residual blocks and skip connections are also highlighted in the works of Amirkolaee and Arefi [36] and Liu et al. [37], who also use an encoder-decoder architecture for their IM2ELEVATION model. Liu et al. additionally apply data preprocessing and registration based on mutual information between the optical image and the DSM. Furthermore, multi-task training proves to be beneficial, especially when depth estimation is combined with image segmentation. Srivastava et al. [38] propose joint height estimation and semantic labelling of monocular aerial images with convolutional neural networks (CNNs). Carvalho et al. [39] use multi-task learning for both the heightmap and the semantics of aerial images. A different approach to the height estimation problem uses a generative model that produces the heightmap of an aerial image given the image as input. This strategy employs the conditional setting of the generative adversarial network and performs image-to-image translation, i.e., the model translates an aerial image to its heightmap. Ghamisi and Yokoya [40] use this exact approach for their IMG2DSM model. Similarly, Panagiotou et al. [41] estimate the Digital Elevation Models (DEMs) of aerial images.

One seemingly reliable feature in remotely sensed images, on which a depth estimation model can rely, is shadows created by the various structures. The length of a shadow is proportional to the height of the object causing it, so drawing the model's attention to evident shadows seems to be a reasonable strategy. Considering the shadows collectively in an image also puts the task into another perspective, mitigating the scale ambiguity by automatically revealing some of the geometric relations between the objects in the scene. We present a DL model that receives a single aerial image and a binary map of its shadows and predicts its heightmap. The binary shadow map is calculated by a simple and efficient algorithm that applies image operators that reveal the shadows in the images. Then, the shadows map is inserted as an extra information channel to the image, alongside the colour channels. Consequently, the model accepts a four-channel input and predicts a heightmap for the image contained in the first three channels of the input (the colour channels of the image). The fourth channel (the shadow map) is used as auxiliary data to motivate the model to pay attention to the shadows. The shadow map comprises an implied attention mechanism that raises model awareness regarding the contribution of shadows to solving the specific task. This simple idea removes the need for costly preprocessing like in Liu et al. [37] and does not intrude on the conventional feature mapping pipeline because it does not alter the image. The model decides whether to use the extra information (shadow map) at specific regions of the image and/or to solely rely on the intensities and colours of the pixels to infer the depth of a building. The shadows map does not exempt the model from the need to consider the pixel intensities and colour. On the contrary, it acts as an extra channel of information. We further apply some architectural characteristics to the Deep Learning model that improve its performance on the specific task. We will describe the task-focused model architecture in the following sections. Figure 1 demonstrates the simple concept of inserting shadow information at the input of the model.

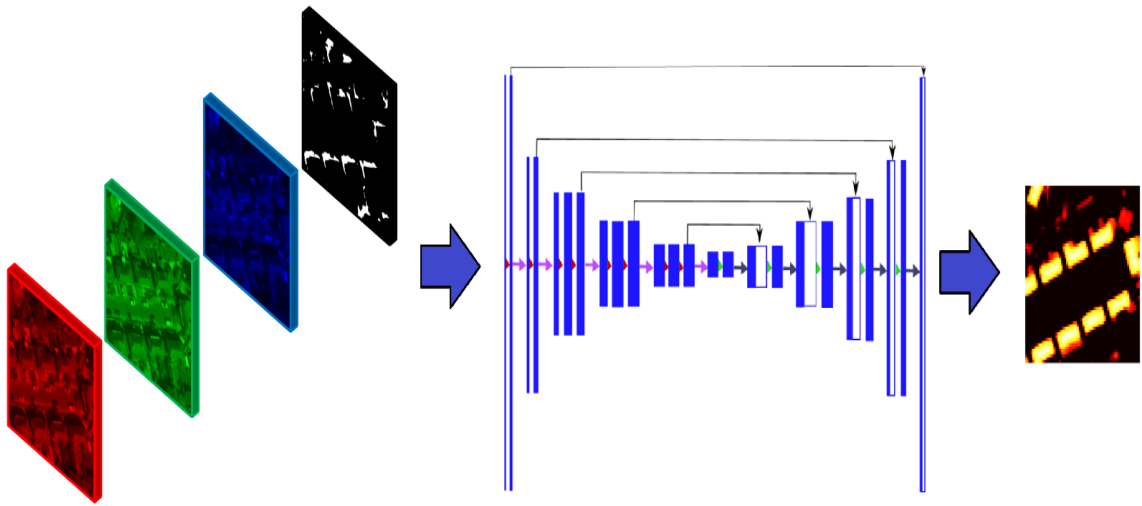

**Figure 1.** Besides introducing some design characteristics that prove to be significantly helpful, we propose feeding the model with an extra channel of information: the shadows map of the image as calculated by a simple and fast algorithm. The shadow map complements the information contained in the colour channels (red, green, blue) of the RGB satellite image and facilitates the heightmap inference process. This technique proves valuable for reducing the prediction error and the number of fail-cases regarding extreme height values that height prediction models usually experience.

Using shadows for predicting the height of buildings has been used before in an analytical and geometric framework where the predictions are based on trigonometric properties [42,43]. These methods utilize both spatial and spectral features from satellite images to enhance shadows and reduce the heterogeneity of intensity to create a contour model. Furthermore, these methods use morphological operations to remove artefacts and require the solar elevation angle to deduce the height of a structure based on the length of its shadow. It is evident that such methods depend on the characteristics of the images and require numerous processing steps to come up with an estimation. They are also limited by their lack of one-step inference, like the case of DL models.

Our model does not require a highly accurate shadows map. Experiments with our model showed that indicating the location of a shadow is the critical factor rather than providing an exact and accurate shadow contour. The model does not significantly benefit from getting accurate shadow maps. It benefits for the cases that it is uncertain of its prediction, and the provided fact that a shadow exists at some specific region of the image enables it to come up with better estimations. In many cases, the model makes appropriate depth estimations regardless of the accuracy of the shadow map.

## 2. Materials and Methods

This section discusses technical aspects of the methods and techniques used in the Deep Learning model. The model's architecture is also presented, with the task-specific design features that make it appropriate for depth prediction. Datasets and the training details are also discussed. For further information on neural networks and Deep Learning, please refer to [44,45].

*2.1 Convolutional Neural Networks*

In contrast to fully connected neural networks where each neuron uses matrix multiplication as its central processing mechanism, CNNs apply convolutions to perform their computations. Each convolutional layer applies several trainable convolutional kernels at its input, resulting in learning appropriate features to tackle a specific task. Convolutional neural networks apply three fundamental concepts in machine learning which comprise the main reasons for their efficiency: sparse interactions between neurons, equivariant representations and parameter sharing. Sparse interactions result from using small kernel sizes compared to the input size and from the fact that deeper layers interact with the input indirectly. Equivariant representations result from sliding a specific kernel on the input, which means that the kernel may detect an exact feature regardless of its position in the input. A convolution can detect a feature in the input even if the input is translated, which means that the operation of convolution has the property of equivariance in translation. Parameter sharing is another essential characteristic of CNNs because it reduces the parameters of the model several orders of magnitude compared to fully connected networks. CNNs are exceptionally efficient when working with structured data that contains spatial relations, such as images. Frequently, in classification tasks, CNNs are used as feature extractors for structured data, and their computations are fed to a classifier that infers the prediction. They are also used for computing meaningful representations that are further analyzed by another model (possibly a second CNN) that infers data for solving other tasks like semantic segmentation [46], pose estimation [47], object localization [48], or auxiliary maps defining an alternative data view. Models that predict a 2D grid are usually fully convolutional and thus contain no fully connected layers.

## 2.2 Residual Networks

The evolution of CNNs through the recent years and their tremendous success in solving complex tasks revealed that the deeper a model is, the better the learning of complex tasks [49]. As much as this was an anticipated fact (since deeper models should produce better and more elaborate representations), in practice, the gradients' vanishing problem prevented the researchers from training very deep models effectively. One of the solutions to the vanishing gradients problem is the use of residual blocks [21]. Instead of having a single forwarding path, an alternative shortcut is available at the input of every block that allows computation to skip some nearby upcoming processing steps. This technique also applies to the reverse direction during backpropagation and allows the gradients to reach more easily and unattenuated the initial layers of the model. A single residual block is shown in Figure 2. Residual neural networks are created by stacking many of these blocks.

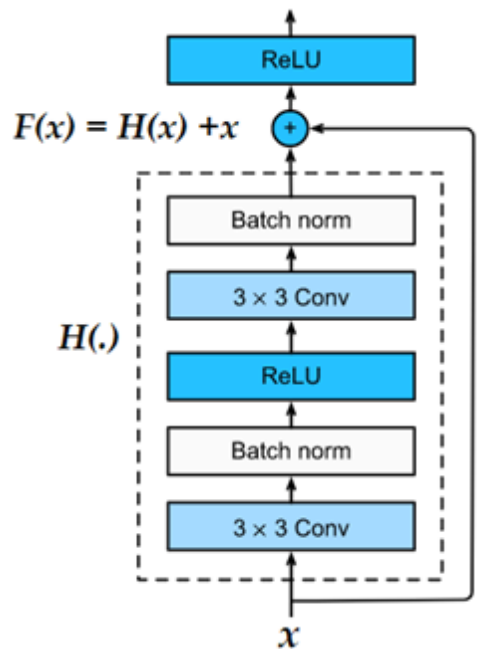

**Figure 2**. The residual block is the building unit of residual neural networks. Input $x$ follows two paths: one that skips some processing steps represented by $H(.)$ and then is added to the result of the operation $H(x)$, and one that applies $H(.)$. This configuration is equivalent to residual learning, and its main advantage is about mitigating the vanishing gradients problem, which is a major problem in training deep neural networks.

An alternative view of residual neural networks is the following: Instead of learning the mapping $F(x)$, we learn the residual mapping $H(x) = F(x) - x$. This enables the model to use the skip connections and the convolutional paths to fit the task best.

## 2.3 The U-Net model

The U-Net model [50] contains an encoding part, a decoding part and some long skip connections that transfer information from the encoder's layers to the decoder's layers. The encoder consistently squeezes the width and height of the feature maps produced by layered convolutions up to the last layer of the encoder, which comprises the middle point of the model. Then, the decoder gradually expands the feature maps until they reach the output size, which is equal to the input size or scaled down by a specific factor. The characteristic U-shape of the model, which explains how the model was attributed its name, comes from the skip connections between the encoder and the decoder. These skip connections allow for the concatenation of feature maps from the encoder's layers with feature maps residing in the decoder's layers and contribute to the recovery of fine-grained detail at the output of the model [51]. Generally, short-skip connections like the ones found in the residual blocks of Residual Networks (ResNets) [21] primarily aim at mitigating the vanishing gradients problem. Long-skip connections additionally aim at recovering spatial information lost during down-sampling. Li et al. [52] show that long-skip connections contribute to a smoother error surface and make training easier and more efficient. Figure 3 shows the general architecture of the U-Net model.

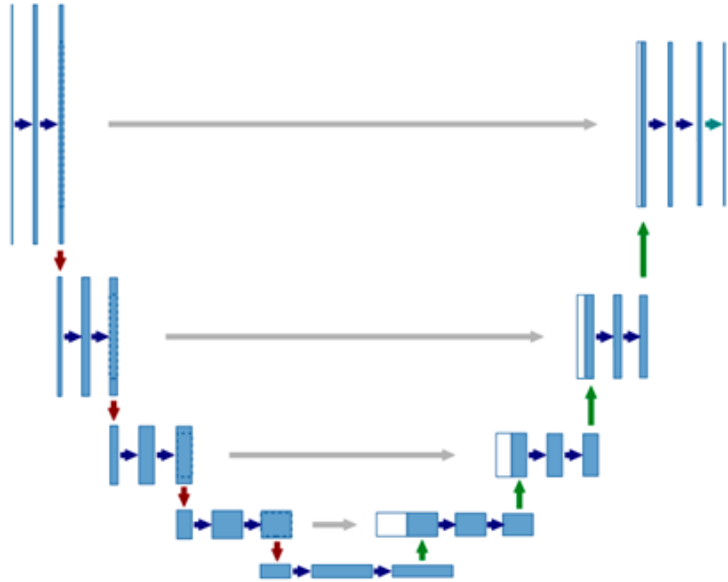

**Figure 3.** The scheme of the U-Net architecture. The left part of the model comprises an encoder that compresses the input to a set of representations. Then, these representations are expanded by the decoder on the right part of the model until they reach an appropriate output size that depends on the task. Skip connections carry spatial information from the early layers (encoder) to the layers of the decoder. The U-Net model shows significant performance, especially on segmentation tasks.

*2.4 Datasets and data pre-processing*

Two relatively large datasets are used to develop and evaluate the proposed height prediction model. The focus of the Manchester area dataset is on estimating the height of buildings, while the focus of the IEEE GRSS data fusion contest dataset is on estimating the heightmaps for all objects in the images.

*2.4.1 Manchester Area Dataset*

The first dataset used for training our model comprises images and Lidar maps from the Trafford area of Manchester, UK. The aerial photography is from Digimap [53] (operated by EDINA [54]), and the RGB images are geospatially aligned according to the UK Grid Reference system. The Lidar data belongs to the UK Environment Agency [55]. It covers approximately 130 $km^2$ comprising roughly 8000 buildings. The RGB images have a resolution of 0.25 $m$, and the Lidar resolution is 1 $m$. The RGB images and the Lidar maps were acquired at different dates; hence there are data inconsistencies like new constructions or buildings demolished. Such inconsistencies constitute a barrier to the training of a DL model. Due to the low Lidar resolution, this dataset is not appropriate for estimating the height of vegetation; thus we concentrate on predicting the height of buildings only. Since we do not have segmentation labels to discriminate what is vegetation and what is not, we apply the simple rule of ignoring heights below 1.5 $m$. This approach affects low vegetation and cars, which is desirable since the cars are mobile objects and thus the source of additional inconsistency. The ground truth heightmaps are computed by subtracting the DTM of each RGB image from its DSM, thus ignoring its elevation. Figure 4 shows examples of different areas from the Manchester dataset and their corresponding ground truth heightmaps. Figure 5 shows the DTM and DSM of Figure 4 (bottom image) and demonstrates some flaws in the specific dataset.

*2.4.2 IEEE GRSS Data Fusion Contest Dataset*

The Data Fusion 2018 Contest Dataset (DFC2018) [56,57] is part of a set of community data provided by the IEEE Geoscience and Remote Sensing Society (GRSS). We specifically use the Multispectral Lidar Classification Challenge data. The RGB images in the dataset have a 5 $cm$ resolution, and the Lidar resolution is 50 $cm$. The data belongs to a $4172 \times 1202$ $m^2$ area. Given the high resolution of the RGB images, this dataset is more suitable than the Manchester area dataset for vegetation detection. Figure 6 shows example pairs of RGB images and their corresponding heightmaps. The ratio between the resolution of the RGB images and the resolution of their corresponding Lidar scans affects the design of the depth-predicting model. Like in the Manchester area dataset, the model must handle the resolution difference between its input and its output and predict a heightmap that is several times smaller than the RGB image at its input. Since the two datasets have different resolutions between the RGB images and the Lidar scans, we cannot use the same model for both cases. Consequently, the models differ in their input/output size and the resolution reduction they must apply to calculate the heightmap. For the most part, the models we use for the two datasets are identical, but we apply slight architectural modifications to cope with the resolution difference.

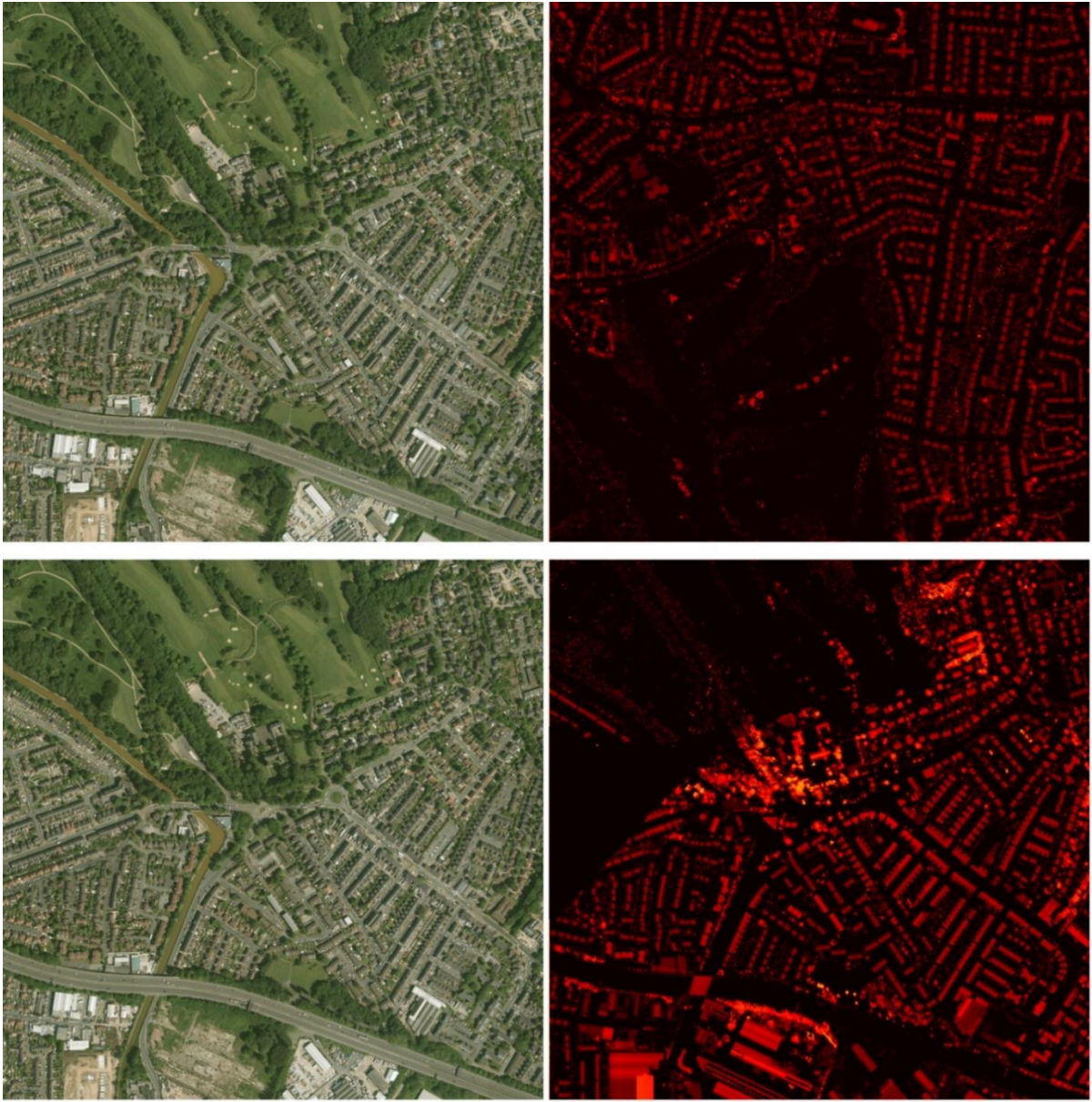

**Figure 4.** Aerial images from the first dataset and their corresponding heightmaps in heat-map format. The aerial images on the left of each pair have a size of 4000 × 4000, while the size of the heightmaps is 1000 × 1000. The heightmaps are created by subtracting the DTM of each image from its DSM. This way, we ignore the altitude information and concentrate on the height of the objects. Furthermore, the heightmaps are shown at the same scale as the satellite images for demonstration reasons. The Figure is best seen in colour.

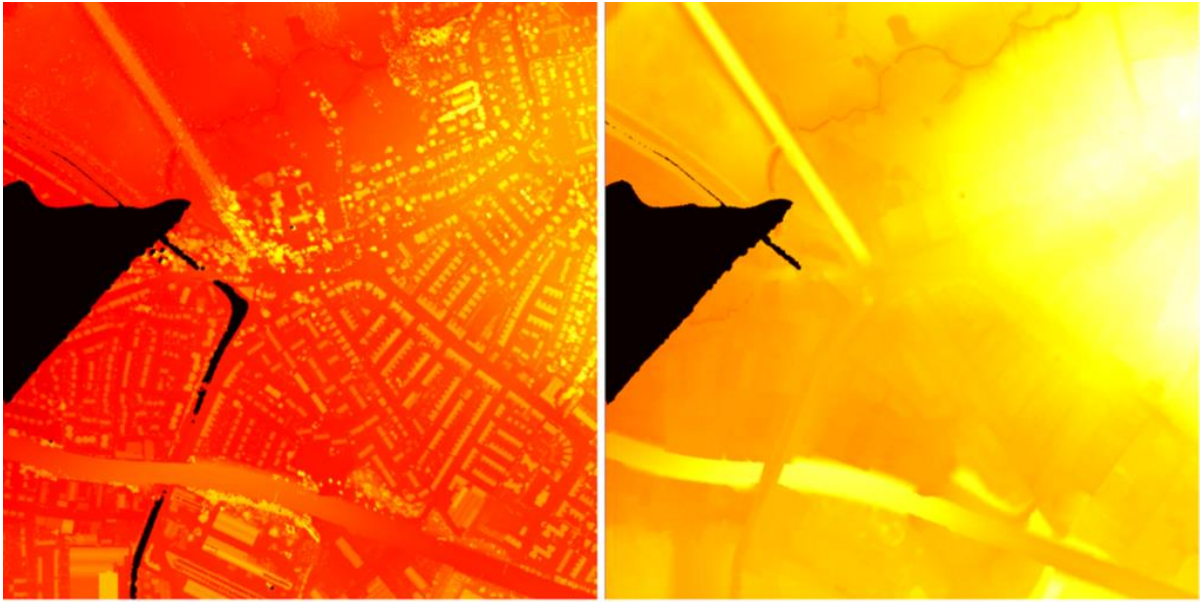

**Figure 5.** The DSM (left) and the DTM (right) of the bottom aerial image of Figure 4. Both maps have several undetermined or irrational (extremely high or low) values shown in black colour. Notably, some of these unexpected values in the DSM map (left) correspond to a river which illustrates a well-known problem of Lidar measurements near highly reflective surfaces. Such erroneous values raise significant problems regarding the training of the model. Thus, they are detected during data preprocessing and excluded from the training data (see Section 2.5). They are also excluded from the validation and test data to avoid false performance evaluation. Overall, these values roughly comprise 10% of the dataset, but they cause larger data discarding since any candidate patch containing even a pixel of undetermined or irrational value is excluded from the training pipeline. This figure is best seen in colour.

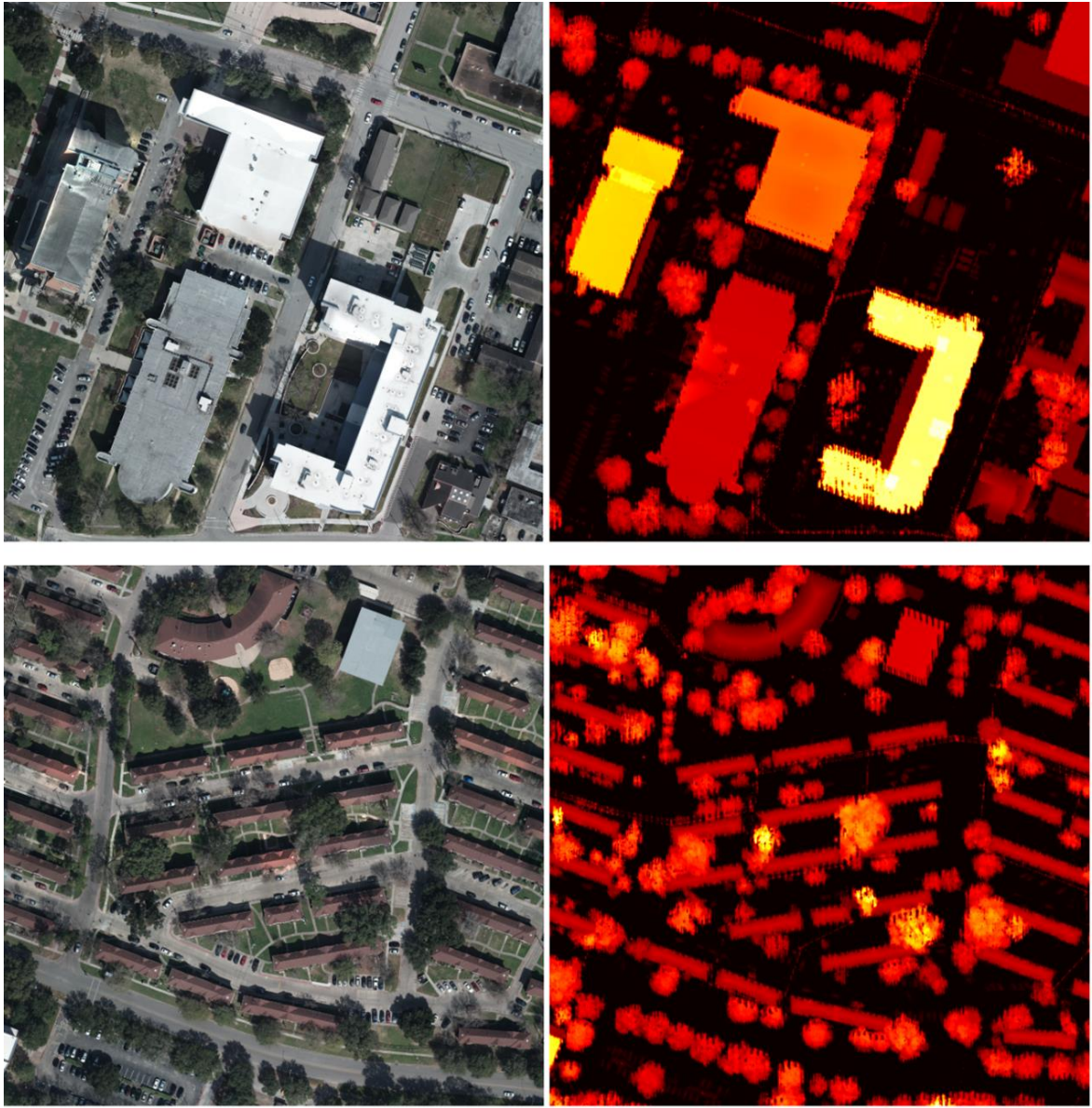

**Figure 6.** Aerial images from the IEEE GRSS Data Fusion Contest (second dataset) and their corresponding height heat maps. The RGB images on the left of each pair have a size of $5000 \times 5000$, while the size of the heightmaps is $500 \times 500$. The heightmaps are created by subtracting the DTM map of each image from its DSM map to ignore the altitude information. The heightmaps are shown at the same scale as the aerial images for demonstration reasons. This figure is best seen in colour.

*2.5 Data Preparation*

The model operating on the Manchester area dataset uses image patches of size $256 \times 256 \times 3$, while the model operating on the DFC2018 dataset uses image patches of size

$520 \times 520 \times 3$. Consequently, the former model outputs a heightmap of size $64 \times 64$, and the latter has an output of $52 \times 52$. These configurations preserve the resolution ratios between the RGB images and the Lidar scans of each dataset. Since our approach aims to make the models focus on the shadows in the aerial images to predict the elevation value for each pixel, we feed the models with appropriate information besides the RGB images. Specifically, we add an information channel for each RGB image that contains a binary map of the shadows. This results in having a model input shape of $256 \times 256 \times 4$ for the first dataset and $50 \times 520 \times 4$ for the second dataset. The last dimension contains the three-colour channels and the shadow map. To obtain the shadow map for each image, some simple image processing operations as used, shown in Table 1. We first increase the contrast of the RGB aerial image, then convert it to a grayscale image, apply some Gaussian blurring, and finally apply a threshold on the values to obtain the locations of the darkest pixels in the processed image which correspond to shadows. These steps reduce the risk of misinterpreting small dark objects like black cars as shadows. Figure 7 shows examples of shadow maps obtained from a $256 \times 256 \times 3$ patch of an aerial image of the Manchester dataset using the procedure described in Table 1. The specific procedure for calculating the shadows map of an image is not computationally expensive and can be applied in the batch creation process during model training. This procedure is more practical than creating the shadow maps for all data before the training begins because some data augmentation techniques like rotation and translation affect the shadow maps.

**Table 1.** Pseudocode of the procedure for calculating the shadows map of aerial images.

Function CreateShadowMap(*image*):

*image*: the aerial image for which we want to calculate the shadows map

1: *image* = EnhanceContrast(*image*)
2: *image* = ConvertGrayscale(*image*)
3: *map* = *image* < 15
4: *map* = Convert2Binary(*map*)
5: return *map*

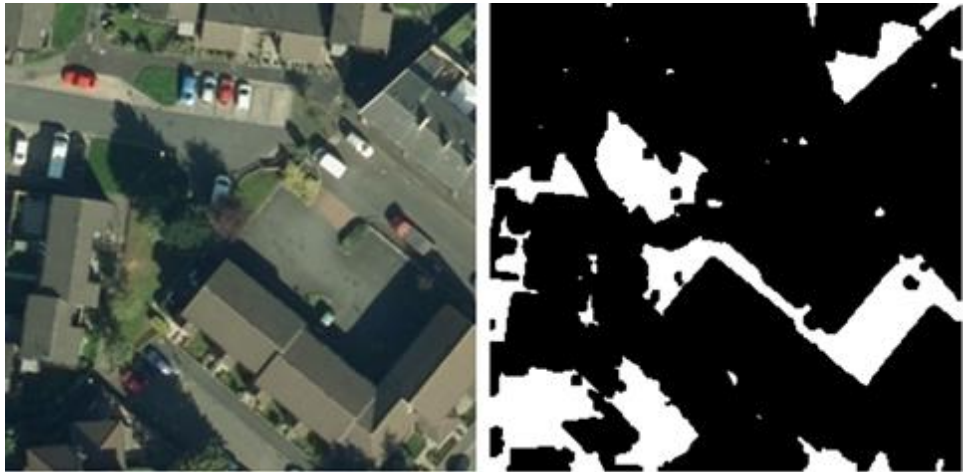

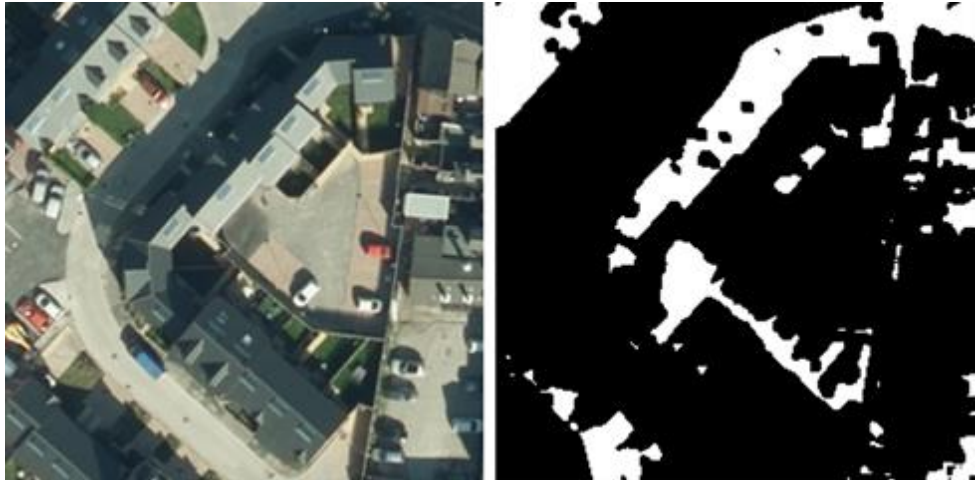

**Figure 7**. Binary shadow maps (at the right of each pair of images) were obtained using the algorithm in Table 1 from patches cropped from RGB images (at the left of each pair). The shadow maps are concatenated with the three colour channels of the RGB images and become a source of additional information for the model, aiming to draw attention to the evident shadows when calculating the heightmap of the patch. The top pair illustrates patches from the Manchester area dataset that have a size of 256 × 256 × 3. The bottom pair contains patches from the second dataset that have a size of 520 × 520 × 3.

*2.6 Model Description*

This section discusses technical aspects of the methods and techniques used in the proposed DL model. The model's architecture is presented, together with the task-specific design features that make it appropriate for depth prediction.

The proposed architecture shares some similarity with semantic segmentation models, where the model must predict the label of each pixel of an image and thus partition it into segments. The segmentation may have the size of the input image, or a scaled-down size. In this study, instead of labels, we predict the real values corresponding to the elevation of each pixel in a down-scaled version of each RGB image. A Lidar instrument provides the ground truth values for these measurements. Like in the semantic segmentation task, several DL models are suitable for learning the task of predicting the Lidar measurements. We choose a popular DL architecture, the U-Net model [50], mainly for its ability to tackle such tasks.

We implement the U-Net architecture with residual blocks both in the encoder and the decoder mechanisms. Specifically, we use three types of residual blocks, as shown in Figure 8:

- A typical residual block (RBLK),
- A down-sampling residual block (DRBLK),
- An up-sampling residual block (URBLK).

A typical residual block contains two convolutional layers at the data path and a convolutional layer with a kernel size of one at the residual connection path. The down-sampling residual block differs in the stride used for the first convolutional layer and the skip connection. Using a higher stride in these convolutions, the previous feature maps are downscaled by a factor $s$ (having a value of 2 in our implementation) in the first convolutional layer of the block and the skip connection, which results in smaller feature maps. The up-sampling residual block uses sub-pixel convolutional up-scaling [58] in the first layer of the block. Sub-pixel upscaling is performed in two steps, with the first step calculating a representation comprising feature maps of size $h \times w \times s^2 c$, where $s$ is the up-scaling factor, and $h \times w \times c$ is the size of the input feature maps. The second step of the process applies a *reshape* operation on the feature maps and produces a representation containing feature maps of size $2h \times 2w \times c$. The skip connection of the up-scaling residual block also applies sub-pixel up-scaling. The detailed architecture of the residual blocks is shown in Figure 8.

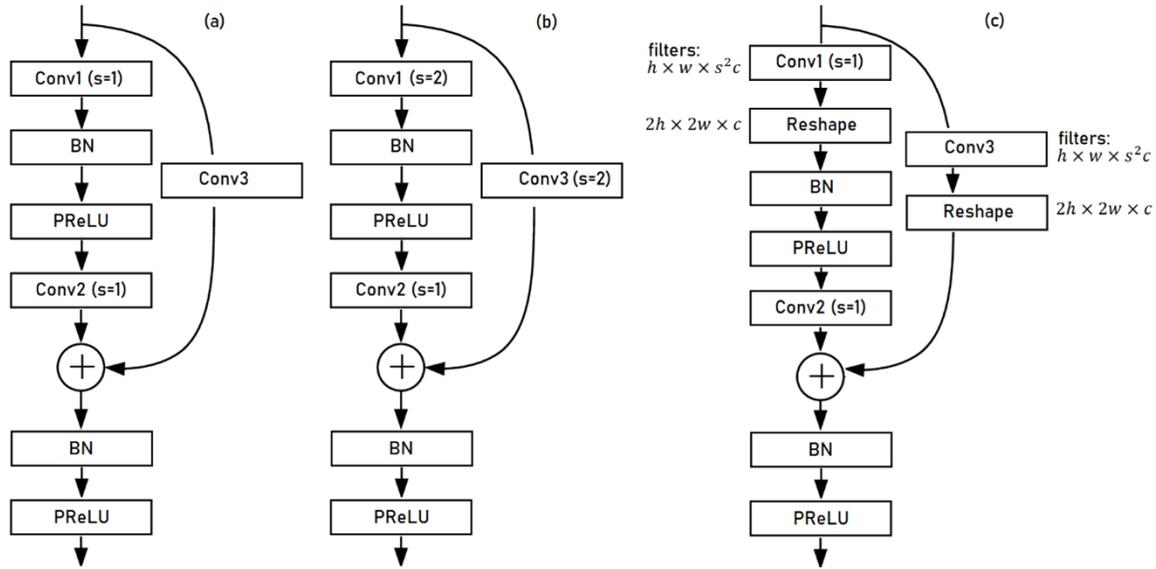

**Figure 8.** The architecture of the three types of residual blocks used in the heightmap prediction model: (a) The typical residual block (RBLK) (b) The down-sampling residual block (DRBLK) uses a stride of two for both the first convolutional layer and the skip layer. (c) The up-sampling residual block (URBLK) uses sub-pixel up-scaling in the first convolution and the skip connection. BN stands for batch normalization [59], PReLU for parametric ReLU, and $s$ for the stride of the convolutional layer.

We use a very similar model for both datasets with minor changes regarding the input patch size and the size of the output heightmap. Furthermore, the Manchester area dataset has an RGB image over depth map resolution ratio equal to 4, so the neural network dealing with this dataset reduces the input size from $256 \times 256 \times 4$ to output size $64 \times 64$. On the other hand, the DFC2018 dataset has an RGB image over depth map resolution ratio equal to 10, so the neural network dealing with this dataset reduces the input size from $520 \times 520 \times 4$ to output size $52 \times 52$. The patch size selected for training the model dealing with the DFC2018 dataset ($520 \times 520 \times 3$) is a compromise between having a manageable input size in terms of memory requirements and having a sufficient heightmap size ($52 \times 52$).

The model dealing with the Manchester area dataset has 164 layers (including concatenation layers and residual blocks' addition layers) and roughly consists of 125 $M$ trainable parameters. Table 2 shows the details of the model architecture. The model dealing with the DFC2018 dataset has 186 layers (including concatenation layers and residual blocks' addition layers) but has fewer parameters to handle the higher memory requirements during training due to the larger input size. Precisely, it consists of 104 $M$ trainable parameters. The only differences with the model used for the Manchester area dataset are a) the addition of some convolutional layers with "valid" padding to achieve the correct output size and b) the reduction of the parameters of the convolutional layers. Table 3 shows the details of the model architecture.

*2.7 Training details*

We apply simple augmentations on the patches during training: rotations of 90, 180 and 270 degrees, small value colour shifting and contrast variations. We also ignore in both datasets patches whose elevation maps contain incomplete or extreme elevation values ($> 100\ m$). Moreover, specifically for the Manchester area dataset, small elevation values ($< 1.5\ m$) were replaced with zero to prevent the model from considering non-stationary objects and low vegetation. This pre-processing is important because of the time difference between the acquisition of the RGB and the Lidar images, which results in inconsistencies between the images and the elevation maps due to the presence of mobile objects like cars in the viewing field of either of the two sensors (RGB or Lidar). The time of acquisition inconsistency in the Manchester area dataset also introduces inconsistencies in vegetation height, and occasionally, in building heights (demolished buildings or new buildings constructed). Consequently, regarding the Manchester area dataset, we focus our attention on predicting the elevation of human-built structures like houses, factories, and public buildings. The DFC2018 dataset has better resolution, and no inconsistencies have been observed. This fact facilitates the prediction of vegetation height as well; thus we do not apply a threshold on the ground truth heightmaps.

The models are trained with the Adam optimizer [60] and a learning rate of $1 \times 10^{-4}$ which decreases the error plateaus for several iterations. Both datasets are randomly split into three sets: a training set (70%), a validation set (15%), and a test set (15%). We use the validation set for hyper-parameter fitting, and we assess the model's performance on the test set. The Mean Absolute Error (MAE) is used between the ground-truth elevation maps and the predicted output as the loss function during training. Mean Squared Error (MSE) was also considered, but it was found that MAE performs slightly better probably because it does not penalize outliers as much as the MSE. We also report the Root Mean Squared Error (RMSE) performance as an additional evaluation metric. All parameters are initialized with the He normal technique [61].

**Table 2.** The model's architecture for the prediction of heightmaps for the Manchester area dataset.

| Layer [1,2] | Output Size (map height/width, channels) | Input |
|---|---|---|
| BN1 [3] | (256,-) | X (256,256,4) |
| Conv1 | (256,64) | BN1 |
| BN2/PReLU1 | (256,-) | Conv1 |
| RBLK1 | (256,64) | BN2/PReLU1 |
| DRBLK1 | (128,64) | RBLK1 |
| RBLK2 | (128,64) | DRBLK1 |
| DRBLK2 | (64,128) | RBLK2 |
| RBLK3 | (64,128) | DRBLK2 |
| DRBLK3 | (32,256) | RBLK3 |
| RBLK4 | (32,256) | DRBLK3 |
| DRBLK4 | (16,512) | RBLK4 |
| RBLK5 | (16,512) | DRBLK4 |
| RBLK6 | (16,1024) | RBLK5 |
| RBLK7 | (16,1024) | RBLK6 |
| URBLK1 | (32,512) | [RBLK7,RBLK5] [4] |
| RBLK8 | (32,512) | URBLK1 |
| URBLK2 | (64,256) | [RBLK8,RBLK4] [4] |
| RBLK9 | (64,256) | URBLK2 |
| RBLK10 | (64,128) | [RBLK9,RBLK3] [4] |
| RBLK11 | (64,128) | RBLK10 |
| RBLK12 | (64,64) | RBLK11 |
| Conv2 | (64,64) | RBLK12 |
| BN3/PReLU2 | (64,-) | Conv2 |
| Conv3 | (64,1) | BN3/PreLU2 |

[1] Convolutional layers not belonging to a residual block (Conv1, Conv2) use "same" padding.
[2] All Convolutional layers use a kernel of size 3.
[3] BN stands for Batch Normalization.
3  [a,b] stands for the concatenation of tensors a and b.

**Table 3.** The model's architecture for the prediction of heightmaps for the DFC2018 dataset.

| Layer [1] | Output Size (map height/width, channels) | Input |
|---|---|---|
| BN1 [2] | (520,-) | X (520,520,4) |
| Conv1 | (256,64) | BN1 |
| BN2/PReLU1 | (256,-) | Conv1 |
| RBLK1 | (256,64) | BN2/PReLU1 |
| DRBLK1 | (128,64) | RBLK1 |
| RBLK2 | (128,64) | DRBLK1 |
| DRBLK2 | (64,128) | RBLK2 |
| RBLK3 | (64,128) | DRBLK2 |
| DRBLK3 | (32,192) | RBLK3 |
| RBLK4 | (32,192) | DRBLK3 |
| DRBLK4 | (16,256) | RBLK4 |
| RBLK5 | (16,256) | DRBLK4 |
| RBLK6 | (16,256) | RBLK5 |
| RBLK7 | (16,512) | RBLK6 |
| URBLK1 | (32,256) | [RBLK7,RBLK5] [3] |
| RBLK8 | (32,256) | URBLK1 |
| URBLK2 | (64,192) | [RBLK8,RBLK4] |
| RBLK9 | (64,128) | URBLK2 |
| RBLK10 | (64,64) | [RBLK9,RBLK3] |
| RBLK11 | (64,64) | RBLK10 |
| Conv2 [4] | (60,64) | RBLK11 |
| BN3/PReLU2 | (60,-) | Conv2 |
| Conv3 [4] | (56,64) | BN3/PReLU2 |
| BN4/PReLU3 | (56,-) | Conv3 |
| Conv4 [5] | (54,32) | BN4/PReLU3 |
| BN5/PReLU4 | (54,-) | Conv4 |
| Conv5 [5] | (52,1) | BN5/PReLU4 |

[1] Convolutional layers not belonging to a residual block (Conv1, Conv2, …, Conv5) use "valid" padding.
[2] BN stands for Batch Normalization.
[3] [a,b] stands for the concatenation of tensors a and b.
[4] Kernel size is 5
[5] Kernel size is 3

## 3. Results

Our model achieves a MAE of 0.52 and an RMSE of 1.22 for the Manchester area dataset, as well as a MAE of 0.71 and an RMSE of 1.39 for the DFC2018 dataset. The lower error values on the first dataset most likely occur due to ignoring small Lidar values, making the model more accurate in predicting buildings and human-made structures. We also train our model without the shadows map being inserted as an additional channel at the input to obtain a baseline for assessing the advantages of using this extra channel of information. The proposed architecture improves the results of Carvallo et al. [39] and Liu et al. [37] by a significant margin (see Table 4), regardless of whether the shadow maps are used or not. Using the additional shadows' channel further reduces the MAE and, more importantly, reduces the RMSE which can be translated as less variance at the residuals. Having less variance implies that using the shadows map at the input is beneficial for reducing the large errors in height-predicting rather than significantly reducing the MAE of the model. The shadows allow the model to identify difficult cases where the RGB information is not informative enough or is misleading the model into making a wrong estimation. The results reveal that the proposed architecture is efficient for the task and achieves significantly better performance *regardless of the additional shadow channel*. Table 4 shows the proposed model's results for both datasets, comparing also with Carvallo et al. [39] and Liu et al. [37].

**Table 4.** Model's performance and comparison to other methods.

| Method | MAE(m) ↓ | RMSE(m) ↓ |
|---|---|---|
| **Manchester Area dataset** [1] | | |
| Our (no shadow channel) | 0.59 | 1.4 |
| Our (with shadow channel) | 0.52 | 1.22 |
| **DFC2018 dataset** [2] | | |
| Carvallo et al. [39] (DSM) | 1.47 | 3.05 |
| Carvallo et al. [39] (DSM + semantic) | 1.26 | 2.60 |
| Liu et al. [37] | 1.19 | 2.88 |
| Our (no shadow channel) | **0.78** | **1.63** |
| Our (with shadow channel) | **0.71** | **1.39** |

[1] 0.25 m/pixel RGB resolution, 1m/pixel Lidar resolution, inconsistencies
[2] 0.05 m/pixel RGB resolution, 0.5m/pixel Lidar resolution

*3.1 Heightmaps estimation for the Manchester Area dataset*

The estimated heightmaps for different areas of the Manchester Area test set are depicted in Figure 9. The estimations are shown in the form of heat maps for better visualization (i.e., providing a more precise display of the relativity between the height values) plus evaluation purposes. Since the model operates on patches of size $256 \times 256 \times 3$, the RGB image is divided into several patches of the specific size and the shadow map for each patch is calculated and added as an extra information channel. Then, the model calculates a high map for each patch and, finally, the estimated maps are recombined to create the overall heightmap for the original RGB image. Interestingly, the model can avoid spiky estimations like the ones indicated as *note 1* in the images of Figure 9: ground truth Lidar maps occasionally contain points of unnaturally high values compared to neighbouring points that constitute false readings happening for several reasons. These reasons relate mainly to the physical properties of the Lidar sensor and the environmental conditions during data acquisition (see Section 1). These wrong readings having values that lie in the high boundary of reasonable Lidar values are very difficult to discriminate from incomplete readings with irrational values. Such spiky readings naturally occur in the training set too. Nevertheless, the model stays unaffected by such inconsistencies in the training set since its estimates regarding spiky measurements occurring at the ground truth data are correct (see Figure 9, *note 1*). Moreover, the Manchester Area dataset contains several inconsistencies in regards to structures that are missing either from the RGB images or from the Lidar maps due to different acquisition times between the two data types. Such inconsistencies are shown in Figure 9 (indicated as *note 2*): In these cases, some structures present in the ground truth heightmap are missing from the RGB images; however, the model correctly predicts the corresponding regions containing the inconsistencies as undeveloped spaces. This behaviour is, of course, desired and proves that our model is robust to false training instances. Furthermore, the results reveal some additional cases which indicate that the model is doing an excellent job estimating the height of buildings, surpassing the quality of the ground truth map. *Note 3* in Figure 9 demonstrates such cases where the ground truth map seems incomplete, spurious and undetailed while the estimation of the model is more detailed and smoother.

*3.2 Heightmaps estimation for the DFC2018 dataset*

The estimated heightmaps for consecutive areas of the DFC2018 test set are illustrated in Figure 10. As in the case of the Manchester Area test set, the RGB image is divided into patches, their shadow map is added as an extra information channel alongside the colour channels, and the model predicts the heightmaps for each of the patches. The only difference is that the size of the patches for this dataset is $520 \times 520 \times 3$ pixels. The estimated heightmaps are then put together to provide the complete heightmap of the entire area. The resulting heightmaps look very similar to the ground truth maps. The higher resolution of the RGB images and the consistency between the RGB and the Lidar measurements in terms of data acquisition time positively impact the model's performance. In this case, the model can estimate vegetation height accurately. Regarding vegetation, the model is consistently overestimating the area covered by foliage, meaning that it fills the space between the foliage. *Note 1* in the second row of Figure 10 (located at the ground truth heightmap) shows the Lidar measurements for an array of trees. Figure 11 shows the magnification of that area in both the ground truth map and the height estimation from the model. This figure demonstrates the tendency of the model to overestimate the volume of foliage. We believe that this behaviour contributes to the higher MAE that the model scores on the DFC2018 dataset compared to the performance on the Manchester area dataset. As mentioned before, the latter dataset has lower resolution and more inconsistencies, but the model training ignores vegetation and low standing objects in its favor. However, we believe that this mistreating could be beneficial under some circumstances, such as projects that focus on vegetation like tree counting, monitoring tree growth or tree coverage in an area [12].

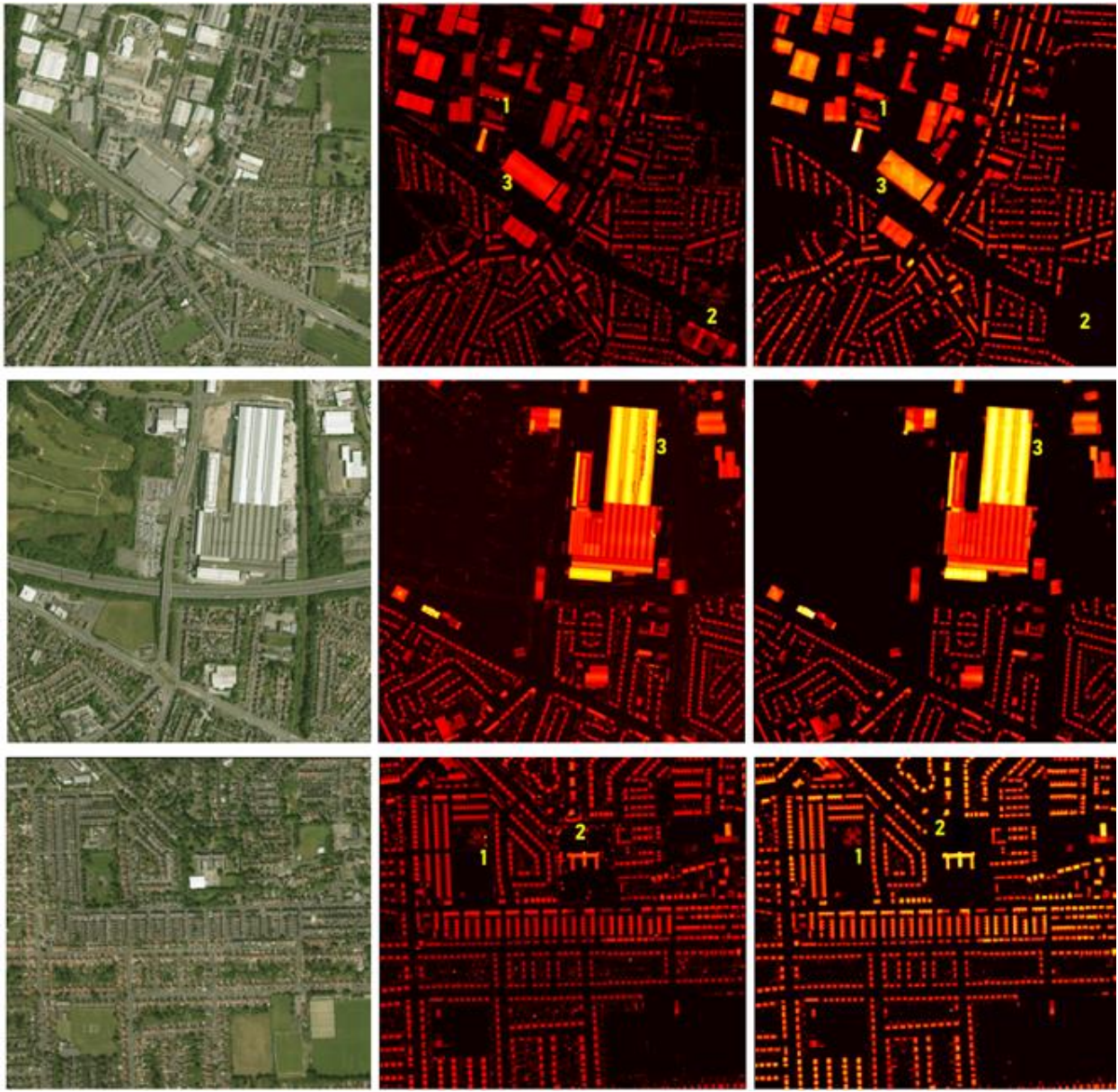

**Figure 9.** Left: Original RGB image of an area in the test set of the Manchester area dataset. Middle: The ground truth heightmap (calculated by DSM-DTM). Right: The elevation heat maps as predicted by the model. *Note 1* shows cases of spurious points in the ground truth that the model correctly avoids estimating. *Note 2* shows occasional inconsistencies in the dataset due to different acquisition of the RGB images and the Lidar measurements. Although these inconsistencies are also evident in the training set, the model is robust to such problematic training instances. *Note 3* shows cases where our model produces better quality maps than the ground truth in terms of surface smoothness and level of detail.

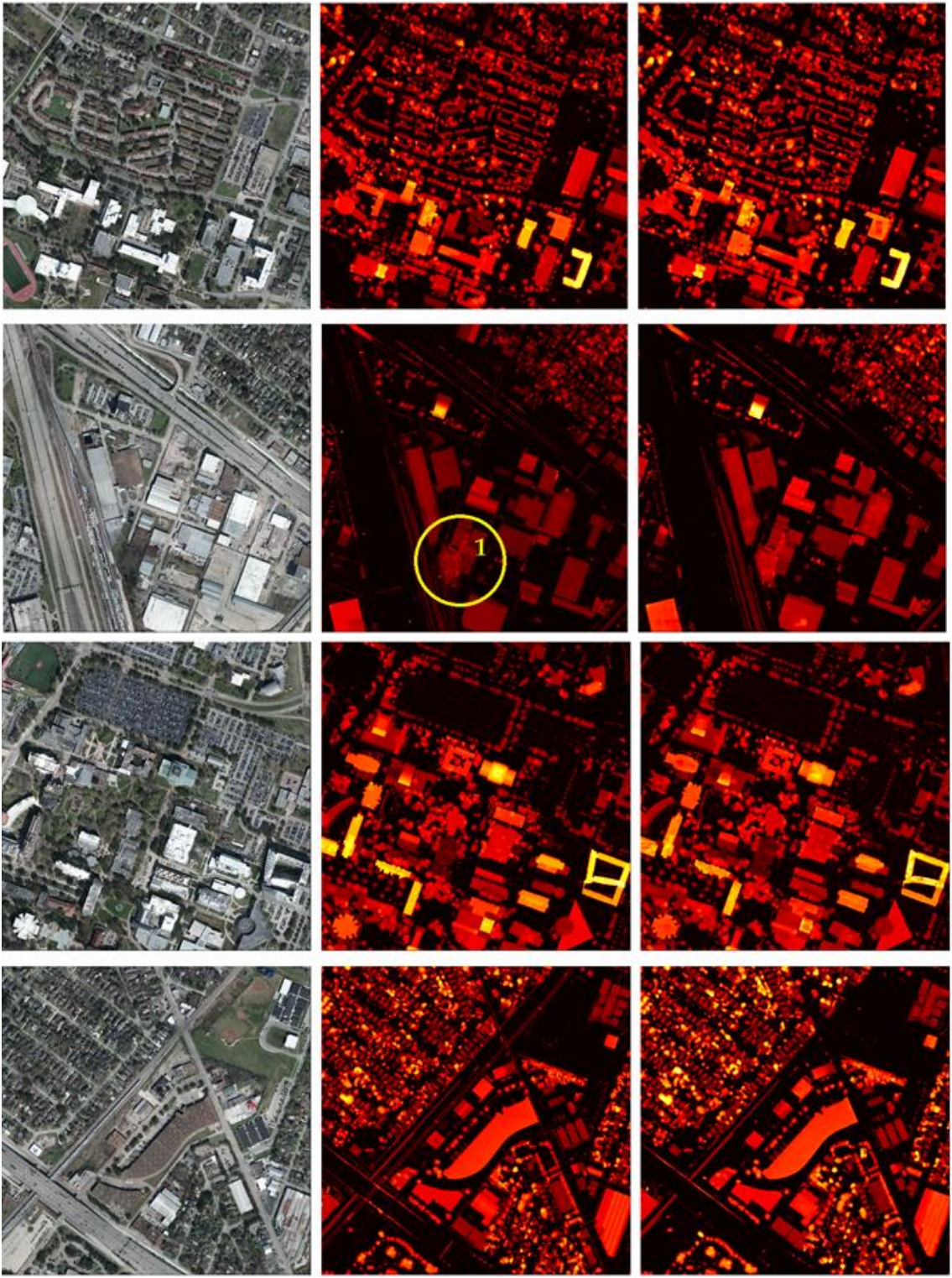

**Figure 10.** Left: Original RGB image from the DFC2018 test set. Middle: Ground truth heightmaps (calculated as DSM-DTM). Right: Model's height estimations. *Note 1* indicates an area that contains an array of trees and is magnified in Figure 11 to demonstrate how the model treats vegetation in the RGB images.

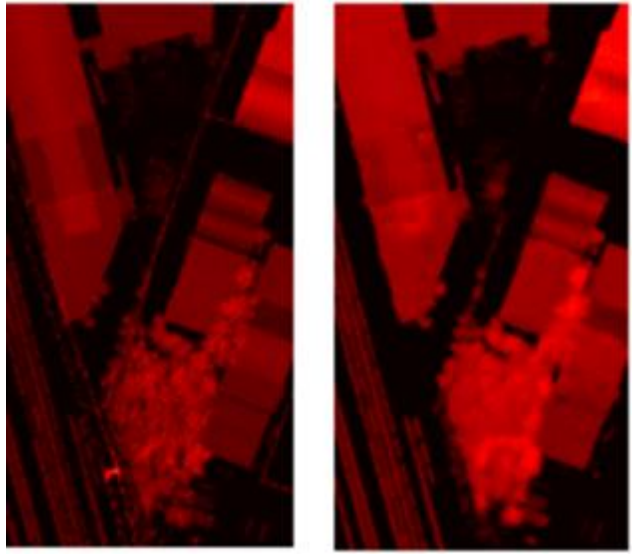

**Figure 11.** Magnification of the noted region (*Note 1*) of Figure 10. Left: Ground truth heightmap. Right: Model output. The model consistently overestimates the foliage volume by filling the spaces between foliage with similar values to neighboring estimations.

*3.3 Investigating the model's reliance on shadows.*

To investigate whether the model is indeed considering shadows for predicting the values of the heightmaps, we conducted further experiments to observe how the model changes its prediction after we manipulate the shadows in the image patch or the shadow map at the input. Specifically, we slide a small square mask over the image patch or the shadow map, which alters the values of the area under the mask. For the RGB images, the sliding window makes the underlying values equal to zero (simulating the presence of a shadow), while for the shadow map, it makes the underlying values equal to one, which again simulates the presence of a shadow in the specific area. In this way, we impose a shadow in specific areas and observe the effect of this modification on the model's prediction (see Figure 12). We also tried removing a shadow from the shadow map or replacing the dark shadow pixels in the images with higher values (thus removing shadows from the image), but this did not influence the prediction as much as adding a shadow. Thus, the effect of manipulating shadows was more evident when adding instead of removing a shadow. This observation implies that the model relies its predictions on both the detection of features of a structure and the shadow that this structure creates. Removing the shadow of a structure results in predicting a slightly lower height for the structure, but this does not fool the model into ignoring the shadowless building. On the contrary, simulating a large shadow anywhere in the image causes the model to increase its elevation prediction for any evident object around that shadow. Figure 12 shows how the model's prediction is affected by sliding a zero-value masking window on the RGB images: a sliding window masks the values in the image that overlap with it, affecting the heat map predicted by the model. When the masking window is nearby structures, it drives the model into predicting larger heigh values for these structures. If no structures exist around the artificial shadow, the impact of the shadow manipulation has less effect on the predicted heightmap or no effect at all.

Summing up, the results of this experiment suggest that the model considers shadows for computing the height of the structures in the RGB image, but it does not solely rely on them. An orphan shadow not associated with any nearby structure does not fool the model into predicting a building nearby. This means that the model combines the presence or even the shape of a shadow, together with the features of a building to compute its height. This experiment also supports the fact that using shadows as an extra information channel at the input does not spectacularly reduce the model loss (MAE). However, it significantly reduces the RMSE of the model (see Table 4), acting as a complementary information source facilitating the height computation.

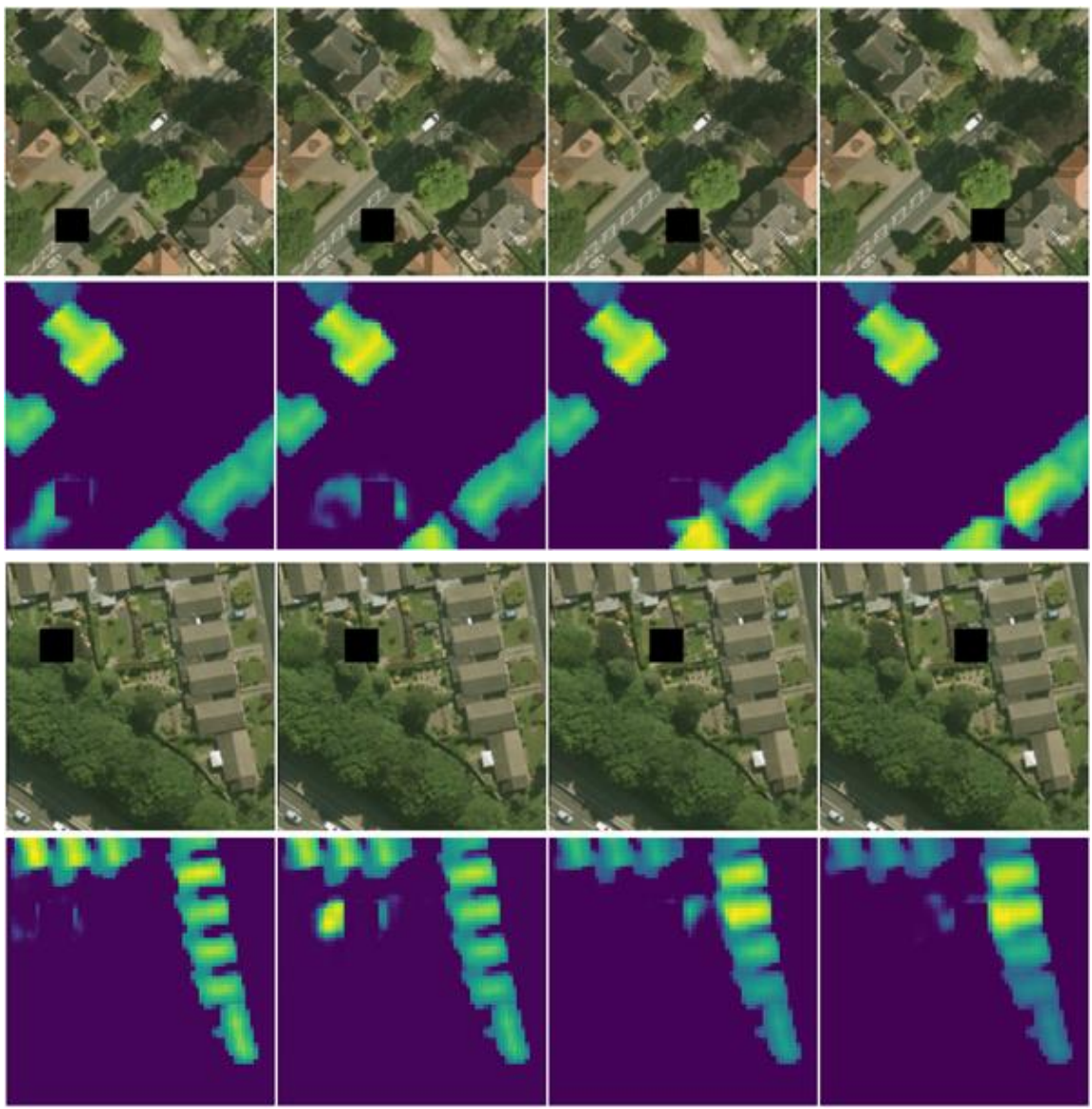

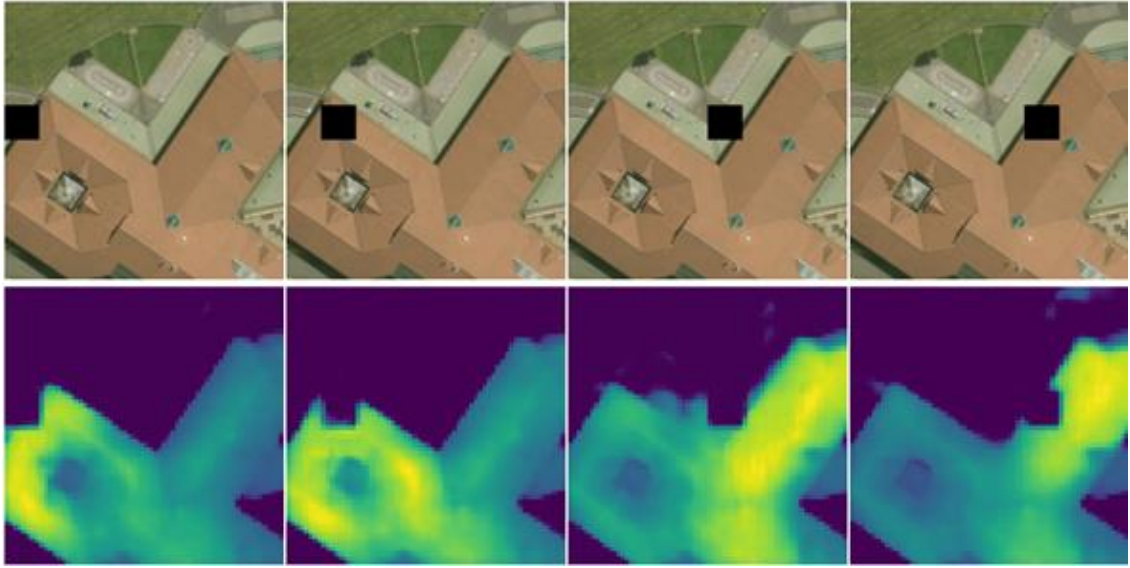

**Figure 12.** Using a sliding masking window to investigate whether the model uses shadows in the objects' height estimation. Every test case is demonstrated in pairs of consecutive rows, the first showing the RGB image and the position of the sliding masking window (black square), while the second row shows the prediction at the output. The artificial shadow implied by the square black box influences the height estimation of the buildings close to the shadow by increasing their height prediction (the values on the predicted map corresponding to the buildings that are close to the shadow become brighter). The estimated height of buildings that are not near the implied shadow is not affected. The artificial shadow causes the model to predict higher elevation for buildings that are in the shadow's proximity. This figure is better seen in colour.

*3.4 Limitations*

Despite the overall promising results of the proposed model, there are still some cases where the model does not perform correctly. Buildings are well represented in both datasets, and thus, the model can predict their height with no problem. The same applies to vegetation in the DFC2018 dataset. However, for objects that are rarely seen in the data (e.g., objects that are tall and thin simultaneously, such as light poles and telecommunication towers), the model has a hard time estimating their height. In cases of such scarce objects, the model treats them as if they do not exist. Rarely seen tall objects that are not bulky or whose structure has empty interior spaces are tough for the model to assess. Examples of such failed cases are shown in Figure 13. The leading cause of the problem is the under-representation of these structures in the dataset. It can be mitigated by introducing more images containing these objects during training.

Although the model is performing well, we acknowledge the fact that it has many parameters. However, predicting the heightmap of a patch is quite fast, especially when the model runs on a Graphical Processing Unit (GPU). Inferring the heightmap of a large area requires the splitting of the RGB image into several patches. Using a GPU, the estimation of the heightmaps for all patches is performed in parallel by processing a batch or batches of patches, taking advantage of the hardware's ability to perform parallel computing.

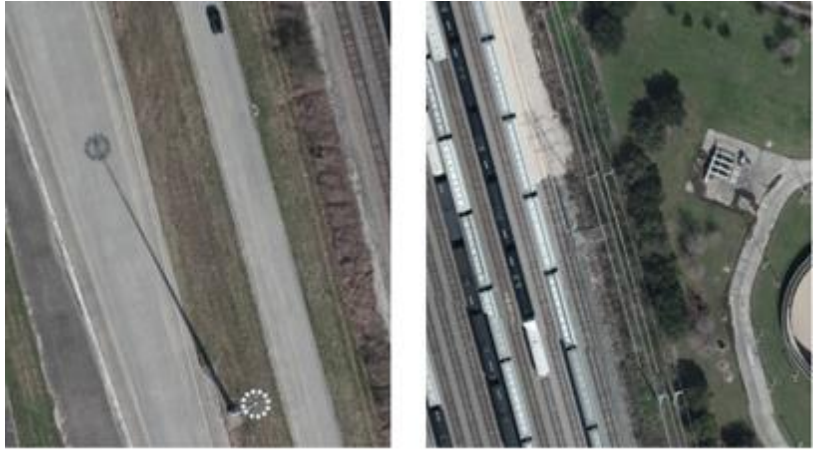

**Figure 13.** Sample failed cases where the model misses the presence of an object completely. The cases are magnified regions from the second RGB image (second row) of Figure 10. The image on the left shows a very high pole standing on a highway with a height of *30* meters (according to its Lidar measurement). Despite the pole's long shadow, the model does not detect it. The magnified region on the right contains a tall electric energy transmission tower that goes undetected by the model.

## 4. Discussion

Obtaining the height of objects in remotely sensed imagery with hardware equipment can be costly, time-consuming, and requires human expertise and sophisticated instruments. Furthermore, the acquisition techniques of such data are demanding and require specialized operators. On the other hand, inferring this data from remotely sensed RGB images is easier, faster, and much less costly. Height estimation from remotely sensed imagery is difficult due to its ill-posed nature, but DL techniques comprise a very promising perspective towards providing adequate solutions to the task.

Our proposed task-focused DL architecture tackles the problem with very good results, which are better than state-of-the-art results in this research problem. A key characteristic of these results is the fact that the variance of the residual error is reduced. We further improve the performance of the proposed DL model by introducing information about the shadows in the image at the input of the model. The model estimates the height of every pixel by considering both the shadow information and the structural features of the objects in the RGB image. We experimented with shadow channel manipulation and have validated the assumption that the model receives information from the shadows around buildings to perform its estimations.

The model has been tested on two different datasets: one with 0.25 $m$ image resolution, 1 $m$ Lidar resolution, and different acquisition times (thus, it has spatial inconsistencies) and one with 0.05 $m$ image resolution and 0.5 $m$ Lidar resolution. The first dataset (capturing lower resolution images) covers the Trafford area in Manchester, UK, while the second dataset is part of the 2018 IEEE GRSS Data Fusion Contest. We use the first dataset to estimate building heights only, while the second dataset is used to estimate both buildings and vegetation heights. Despite the inconsistencies encountered in the first dataset, the effectiveness of the model indicates its high robustness and ability to build domain knowledge without resorting to dataset memorization. This indication is also suggested by the fact that we did not use data curation or any special prepossessing besides data augmentation.

We aspire that the possibility of deriving high-precision digital elevation models from RGB images and shadows alone, without expensive equipment and high costs, will accelerate global efforts in various application domains that require geometric analysis of areas and scenes. Such domains include urban planning and digital twins for smart cities [11], tree growth monitoring and forest mapping [12], modelling ecological and hydrological dynamics [62], detecting farmland infrastructures [63], etc. Such low-cost estimation of buildings' heights will allow policy-makers to understand the potential revenue of rooftop photovoltaics based on yearly access to sunshine [64] and law enforcement to verify whether urban/or rural infrastructures comply with local land registry legislation.

Finally, we note that the model experiences some ill-cases with tall-thin and under-represented objects. This issue can be solved by including more examples of such objects in the training images, which is an aspect of future work.